\documentclass[10pt,twocolumn,letterpaper]{article}

\usepackage{iccv}              

\definecolor{iccvblue}{rgb}{0.21,0.49,0.74}
\usepackage[pagebackref,breaklinks,colorlinks,allcolors=iccvblue]{hyperref}
\usepackage{url}
\usepackage{bm}

\def\paperID{20} 
\def\confName{ICCV}
\def\confYear{2025}

\title{Engagement Prediction of Short Videos with Large Multimodal Models}

\author{Wei Sun$^{1}$\thanks{These authors contributed equally to this work.}, Linhan Cao$^{2}$\footnotemark[1], Yuqin Cao$^{2}$, Weixia Zhang$^{1}$\thanks{Corresponding authors.}, Wen Wen$^{3}$, \\ Kaiwei Zhang$^{2}$, 
Zijian Chen$^{2}$, Fangfang Lu$^{4}$, Xiongkuo Min$^{2}$, Guangtao Zhai$^{2}$\footnotemark[2]\\
$^1$East China Normal University, $^2$Shanghai Jiao Tong University\\
$^3$City University of Hong Kong, $^4$Shanghai University of Electric Power
}

\begin{document}
\maketitle

\begin{abstract}
The rapid proliferation of user-generated content (UGC) on short-form video platforms has made video engagement prediction increasingly important for optimizing recommendation systems and guiding content creation. However, this task remains challenging due to the complex interplay of factors such as semantic content, visual quality, audio characteristics, and user background. Prior studies have leveraged various types of features from different modalities, such as visual quality, semantic content, background sound, etc., but often struggle to effectively model their cross-feature and cross-modality interactions. In this work, we empirically investigate the potential of large multimodal models (LMMs) for video engagement prediction. We adopt two representative LMMs: \textbf{VideoLLaMA2}, which integrates audio, visual, and language modalities, and \textbf{Qwen2.5-VL}, which models only visual and language modalities. Specifically, VideoLLaMA2 jointly processes key video frames, text-based metadata, and background sound, while Qwen2.5-VL utilizes only key video frames and text-based metadata. Trained on the SnapUGC dataset, both models demonstrate competitive performance against state-of-the-art baselines, showcasing the effectiveness of LMMs in engagement prediction. Notably, VideoLLaMA2 consistently outperforms Qwen2.5-VL, highlighting the importance of audio features in engagement prediction. By ensembling two types of models, our method achieves first place in the ICCV VQualA 2025 EVQA-SnapUGC Challenge on short-form video engagement prediction. The code is available at \url{https://github.com/sunwei925/LMM-EVQA.git}.
\end{abstract}

\section{Introduction}

The widespread popularity of short-form video platforms such as TikTok, YouTube Shorts, and Snapchat has led to an unprecedented surge in user-generated content (UGC). With millions of UGC videos uploaded to these platforms every day, accurately predicting video engagement (\textit{i.e.,} estimating how likely a video is to capture and sustain viewer attention) has become essential for optimizing content recommendation, enhancing user retention, and also guiding content creation.

Video engagement is commonly measured using metrics such as view count, average watch time, average watch percentage, and like rate~\cite{bulathwela2020vlengagement}. However, these metrics can be biased; for instance, longer videos are inherently less likely to be watched in full compared to shorter ones. To address such limitations, several studies have proposed more robust engagement metrics. Wu \textit{et al.}~\cite{wu2018beyond} introduce a relative engagement metric that accounts for varying video durations by modeling mutual relationships and ranking orders among videos with similar durations. Zhan \textit{et al.}~\cite{zhan2022deconfounding} suggest training separate models for videos of different durations to mitigate duration-related bias. Li \textit{et al.}~\cite{li2024delving} propose two alternative metrics: Normalized Average Watch Percentage (NAWP), which normalizes average watch percentage across videos, and Engagement Continuation Rate (ECR), which measures the proportion of users who watching beyond a certain duration (\textit{e.g.}, five seconds). In this paper, we focus on the ECR metric, as it better reflects a video's ability to attract and retain viewer attention, and is also highly correlated with NAWP.

However, predicting ECR in cold-start scenarios (\textit{i.e.}, when newly uploaded videos lack any user interaction data) remains a challenging task, as it requires modeling the complex interplay among various factors, including the video's semantic content, visual quality, background audio, and user-related characteristics. Recent studies~\cite{wu2018beyond,li2024delving} have attempted to address this by leveraging various features from multiple modalities, such as video quality assessment (VQA) features~\cite{min2024perceptual}, aesthetic features~\cite{sun2024assessing}, background sound~\cite{cao2025agav}, and textual metadata (\textit{e.g.}, titles and descriptions)~\cite{li2024delving}, video captioning~\cite{li2024delving}, etc., and feeding them into regression networks to estimate the ECR value. Nevertheless, these approaches often suffer from limitations in the expressiveness of the extracted features and the ability of regression networks to effectively model cross-feature and cross-modality interactions, resulting in suboptimal performance.

In addition to the two-stage video engagement prediction paradigm, large multimodal models (LMMs), which directly model multi-modal data such as visual, audio, and textual inputs, hold strong potential for predicting video engagement, yet remain largely unexplored. In this work, we empirically investigate the effectiveness of LMMs for ECR prediction. Given that video engagement is influenced by a combination of factors, we adopt two representative LMMs: \textbf{VideoLLaMA2}~\cite{cheng2024videollama}, an audio-visual-language model, and \textbf{Qwen2.5-VL}~\cite{bai2025qwen2}, a visual-language model—--both state-of-the-art in their respective domains. By comparing these two types of LMMs, we could further analyze the role of audio in video engagement prediction.

Specifically, we categorize the inputs to LMMs into three modalities: (1) \textbf{visual input}, which consists of key video frames used to extract semantic content, visual quality, aesthetic attributes, and other visual features; (2) \textbf{textual metadata}, including video titles and descriptions that provide keywords and high-level summaries of the content (e.g., video category or topic); and (3) \textbf{background sound}, which captures audio information such as music and speech. For VideoLLaMA2, all three modalities are utilized, while Qwen2.5-VL leverages only visual input and textual metadata. In terms of prediction strategy, we explore two types of regression methods. For VideoLLaMA2, we extract the hidden representation from the model’s final layer and append a lightweight regression head composed of a two-layer multilayer perceptron (MLP). In contrast, for Qwen2.5-VL, we formulate ECR prediction as a token-level regression task, where the model directly generates a numerical value as output.

By training both models on the SnapUGC dataset~\cite{li2024delving}—a large-scale collection of user-generated short videos from Snapchat Spotlight annotated with ECR values—we observe that both types of LMMs consistently outperform the baseline model, a two-stage video engagement prediction method. This demonstrates the strong potential of LMMs for video engagement prediction. Furthermore, we find that VideoLLaMA2 significantly outperforms Qwen2.5-VL, despite the latter being a more recent model with superior performance on general visual tasks. This performance gap highlights the important role of audio in engagement prediction and suggests that audio information should not be overlooked. Finally, by ensembling predictions from both models, our method achieves first place in the EVQA-SnapUGC Challenge on short-form video engagement prediction~\cite{li2025evqa} held by the ICCV VQualA 2025~\cite{isrgcq2025iccvw,zhu2025vqa,genai-bench2025iccvw,li2025evqa,ma2025fiqa,diqa2025iccvw}.

In summary, the main contributions of this paper are as follows:
\begin{itemize}
    \item We propose leveraging large multimodal models for end-to-end video engagement prediction using multi-modal inputs, including visual, audio, and textual information.
    
    \item We design and evaluate two representative LMMs: Video-LLaMA2, an audio-visual-language model, and Qwen2.5-VL, a visual-language model. Each employs a different regression strategy for engagement prediction.
    
    \item We validate our approach on the SnapUGC dataset and achieve competitive performance, highlighting the potential of general-purpose LMMs, especially audio-visual-language models, for real-world engagement understanding tasks.
\end{itemize}

\section{Related Work}
We review related work from two aspects: video quality assessment, a closely related field that focuses on evaluating visual quality, and video engagement prediction.

\subsection{Video Quality Assessment}

Video quality assessment (VQA)~\cite{min2024perceptual} aims to automatically evaluate the visual quality of a given video and is a well-established research topic that has been studied for many years. In the literature, visual quality is typically assessed by human viewers and annotated using Mean Opinion Scores (MOS), which represent the average of subjective quality ratings collected from a small group of individuals. In general, constructing such VQA datasets with MOS is referred to as subjective VQA, while developing models to automatically predict quality scores is known as objective VQA.

Early subjective VQA datasets~\cite{lee2021subjective, li2019avc, mackin2015study, madhusudana2021subjective, nasiri2015perceptual, de2010h, seshadrinathan2010study, vu2014vis} primarily target synthetic distortions introduced by various video processing stages, such as compression, transmission, up/down-sampling, etc. These datasets typically consist of a limited number of pristine source videos along with their artificially distorted counterparts. While valuable for benchmarking classical VQA algorithms, they suffer from low content diversity and limited realism, which hinder their generalizability to UGC videos on social media platforms. To address this limitation and enhance real-world applicability, more recent datasets~\cite{ghadiyaram2017capture, ying2021patch, nuutinen2016cvd2014, hosu2017konstanz, yim2020subjective, sinno2018large} have shifted their focus toward authentic distortions captured in the wild, including camera shake, focus issues, exposure imbalance, compression artifacts etc. Furthermore, with the rapid rise of generative models, a new line of research~\cite{li2025aghi,chen2024gaia,chen2024study,zhang2024human,zhang2024benchmarking} has emerged to study distortions introduced by AI-generated content (AIGC) algorithms, which present unique challenges due to their semantic inconsistencies, unnatural textures, and generation-induced artifacts.

Objective VQA algorithms can be broadly categorized into three types: handcrafted feature-based methods, deep neural network (DNN)-based methods, and multimodal-based methods. Specifically, handcrafted feature-based VQA methods~\cite{ebenezer2021chipqa, saad2014blind, mittal2015completely, korhonen2019two, tu2021ugc, tu2021rapique} primarily rely on manually designed features that capture fundamental perceptual cues such as brightness, contrast, edge sharpness, and motion artifacts. A common paradigm involves first extracting quality-relevant features and then applying a regression model, such as a support vector regressor (SVR), to predict the quality score. For example, V-BLIINDS~\cite{saad2014blind} leverages natural scene statistics (NSS) across both spatial and temporal domains and uses SVR for quality prediction. TLVQM~\cite{korhonen2019two} introduces a comprehensive set of spatiotemporal features, including jitter, blur, and noise, combined with SVR and random forest regressors. Although handcrafted features are interpretable, they often struggle to capture the complex and subtle distortions present in real-world videos, limiting the overall effectiveness and generalizability of these models.

DNN-based VQA methods~\cite{li2022blindly, li2019quality, liu2023ada, liu2018end, ying2021patch, sun2021deep, yi2021attention, sun2022deep, sun2023blind, zhang2023md, sun2024analysis, wu2022fast, wu2023exploring, sun2024enhancing,lu2023bh,sun2025empirical,sun2025compressedvqa} leverage deep neural networks, such as convolution neural networks (CNN) and Transformers, to automatically extract quality-aware features. For example, VSFA~\cite{li2019quality} employs a pre-trained CNN to extract frame-level features and uses a GRU network to model temporal dependencies. Li~\textit{et al.}~\cite{li2022blindly} integrate both IQA and action recognition networks to jointly model spatial and temporal aspects of video quality.  SimpleVQA~\cite{sun2022deep} and MinimalisticVQA~\cite{sun2024analysis} design a simple architecture that incorporates trainable spatial extractors and pre-trained motion modules. FAST-VQA~\cite{wu2022fast} samples spatiotemporal grid mini-cubes and utilizes a transformer-based attention mechanism for end-to-end modeling. DOVER~\cite{wu2023exploring} extends FAST-VQA by adding an additional branch to assess aesthetic quality.  Despite these advancements, most existing methods rely solely on the visual modality and overlook critical engagement-related factors such as background music, sound effects, and textual overlays. This unimodal limitation restricts their effectiveness in modeling user engagement in short-form videos, where the interplay between visual, audio, and textual cues is essential.

To overcome this limitation, recent studies~\cite{li2024llava, bai2025qwen2, zhu2025internvl3, wu2024q, cao2025breaking,wu2025fvq,zhang2025leveraging,zhang2025q,zhang2024lmm,wang2024large} have explored the use of large multimodal models (LMMs), which exhibit strong capabilities in understanding and reasoning over both visual and textual inputs, to improve VQA performance. Representative multimodal-based VQA methods such as Q-Align~\cite{wu2023q}, LMM-VQA~\cite{ge2024lmm}, VQA-Scorer~\cite{jia2024vqa}, and LMM-PVQA~\cite{cao2025breaking} align visual encoders with large language models (LLMs) using instruction tuning or alignment-based training objectives. These models leverage natural language queries to interpret low-level visual distortions and show promising performance in zero-shot and open-set quality assessment scenarios. In addition, AGAV-Rater~\cite{cao2025agav} introduces a benchmark for evaluating the perceptual quality of AI-generated audio-visual content. It employs a transformer-based architecture that fuses audio and visual features extracted from pre-trained encoders, followed by a lightweight regression head built on top of a language model to predict quality scores.

\begin{figure*}[t]
    \centering
    \includegraphics[width=1.05\linewidth]{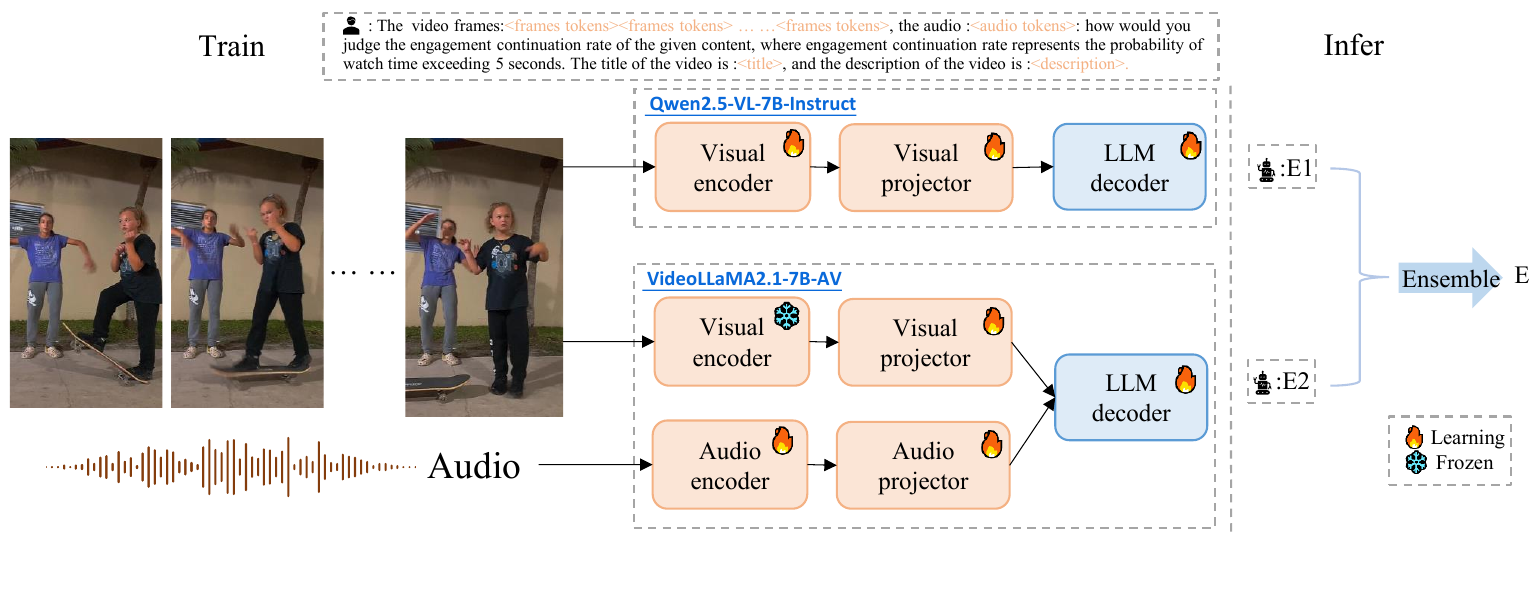}
    \caption{Overview of the proposed ensemble framework for engagement continuation rate prediction. The model integrates two large multimodal branches: Qwen2.5-VL-7B-Instruct and Video-LLaMA2-7B-AV. Each branch takes keyframes and textual metadata as input, while Video-LLaMA2 additionally incorporates audio signals. Visual and audio features are first encoded and projected into the language space, then fused with the textual prompt and processed by the LLM decoder.}
    \label{fig:framework}
\end{figure*}

\subsection{Video Engagement Prediction}
Several recent works~\cite{wu2018beyond,bulathwela2020vlengagement,zhan2022deconfounding,li2024delving} have explored large-scale datasets and models for video engagement estimation under cold-start settings. Wu \textit{et al.}~\cite{wu2018beyond} construct a large-scale dataset of 5.3 million YouTube videos to study video engagement beyond views. They define relative engagement, a duration-normalized watch percentage, and show it is stable over time and correlated with video quality. Using only content and channel features, their model predicts engagement in a cold-start setting with high accuracy, demonstrating the feasibility of pre-upload engagement prediction. Bulathwela \textit{et al.}~\cite{bulathwela2020vlengagement} propose VLEngagement, a large-scale dataset containing 4,046 scientific video lectures annotated with user engagement signals, including average star ratings, view counts, and normalized median and average watch time. The dataset incorporates a wide range of content-based, Wikipedia-based, and video-specific features. Ensemble models such as random forests and gradient boosting machines are used to predict and rank engagement levels, providing strong baselines for context-agnostic engagement prediction. Zhan \textit{et al.}~\cite{zhan2022deconfounding} introduce a large-scale dataset from Kuaishou with over 1.3 billion user-video interactions, labeled with actual watch time and video duration. To address duration bias in watch-time prediction, they propose the D2Q framework, which splits the data by duration and learns a unified quantile regression model for watch-time estimation. Its enhanced version, Res-D2Q, further incorporates duration as a network input, leading to improved performance in both offline and online evaluations. Li \textit{et al.}~\cite{li2024delving} introduce SnapUGC, a large-scale dataset of 90,000 short videos from Snapchat Spotlight, annotated with two engagement metrics—normalized average watch percentage (NAWP) and engagement continuation rate (ECR)—aggregated from over 2,000 real users per video. To predict engagement in a cold-start setting, they propose a multimodal model that integrates visual, audio, and textual features via cross-modal attention, achieving superior performance over traditional VQA-based methods.

\section{Method}

In this section, we introduce the proposed LMM-based model for video engagement prediction in cold-start scenarios. An overview of the model is shown in Figure~\ref{fig:framework}. The LMM takes as input multi-modal signals, including visual inputs, background audio, and textual metadata, and regresses them to an engagement score.

\subsection{Multi-modal Inputs}
To capture the diverse factors influencing user engagement for UGC videos, we incorporate three complementary input modalities: visual, textual, and audio. Each modality contributes information to characterize the semantic content, aesthetic appeal, and contextual cues for video engagement.

\begin{itemize}
    \item \textbf{Visual Input.} We extract a set of keyframes from each video using uniform sampling to capture its visual dynamics and content. Specifically, since we focus on the engagement continuation rate, which measures the proportion of users who watch a video for at least five seconds, we extract the first eight keyframes from each video as the visual input. We denote the set of keyframes as $\bm x = \{\bm x_i\}^{N}_{i=1}$, where $x_i$ represents the $i$-the keyframe and $N=8$.
    \item \textbf{Textual Metadata.} Each video is accompanied by user-provided or platform-generated metadata, including titles and descriptions. These texts offer high-level summaries of the video’s topic, category, or style, and often contain keywords indicative of viewer interest. If either field is missing, we replace it with the placeholder token``None" to keep input format consistency. We tokenize and embed these texts using the respective language encoder in each model to integrate language-level priors. The title and description of a video are denoted as $T_{\text{title}}$ and $T_{\text{description}}$, respectively.
    \item \textbf{Background Audio.} Background audio, such as music, speech, sound effects, and ambient noise, plays a crucial role in influencing viewer engagement. It helps establish the emotional atmosphere, highlight key moments, and deliver narrative cues that enrich the overall viewing experience. Optionally, we incorporate the background audio, denoted as $A$, as an input to the model.
\end{itemize}

\subsection{Model Structures}
We explore two representative LMMs for the video engagement prediction task: VideoLLaMA2~\cite{cheng2024videollama} and Qwen2.5-VL~\cite{bai2025qwen2}. In addition, we equip each model with different regression strategies to investigate their impact on LMM-based video engagement prediction.

\subsubsection{VideoLLaMA2}
VideoLLaMA2 is a large multimodal model designed for audio-visual and spatiotemporal understanding. It is originally designed to take video frames, textual instructions, and audio signals as inputs. Accordingly, we provide VideoLLaMA2 with the visual input $\bm{x}$, textual metadata $T_{\text{title}}$ and $T_{\text{description}}$, and background audio $A$.

The keyframes $\bm{x}$ are first resized to $384 \times 384$ and then encoded into visual embeddings ${\mathbf{z}i^{\text{visual}}}$ using the vision encoder $f{\text{visual}}$:

\begin{equation}
\label{visual_equation}
\mathbf{z}_i^{\text{visual}} = f_{\text{visual}}(\mathbf{x}_i), \quad i=1,\ldots,N.
\end{equation}

The complete audio track is converted into a spectrogram and encoded into a sequence of audio embeddings ${\mathbf{z}^{\text{audio}}}$ using the audio encoder $f_{\text{audio}}$:
\begin{equation}
\mathbf{z}^{\text{audio}} = f_{\text{audio}}(A).
\end{equation}

The visual and audio tokens are then formatted into a natural language prompt $P$ by inserting modality-specific tokens, along with the textual metadata:

\begin{quote}
\small
\texttt{The video frames: <frames tokens> ... <frames tokens>, the audio: <audio tokens>. How would you judge the engagement continuation rate of the given content, where engagement continuation rate represents the probability of watch time exceeding 5 seconds. The title of the video is: <title>, and the description of the video is: <description>.}
\end{quote}

The entire prompt is then tokenized and encoded using the LLM's language encoder $f_{\text{text}}$, resulting in a sequence of token embeddings ${\mathbf{z}_i^{\text{text}}}$:

\begin{equation}
\mathbf{z}^{\text{text}} = f_{\text{text}}(P).
\end{equation}

In contrast to previous methods that discretize the continuous score into categorical classes~\cite{wu2023q}, we formulate the task as a continuous regression problem. Given the fused multimodal input $\mathbf{Z} = [\mathbf{z}^{\text{visual}}, \mathbf{z}^{\text{audio}}, \mathbf{z}^{\text{text}}]$, the LLM decoder produces a sequence of hidden states $\mathbf{h}_{1:T}$. We compute the mean hidden state $\hat{\mathbf{h}}$ and use it to predict a scalar ECR score $\hat{y}$ through a lightweight regression head $g(\cdot)$ implemented via a multi-layer perceptron:
\begin{align}
\hat{\mathbf{h}} &= \frac{1}{T}\sum_{i=1}^{T}\mathbf{h}_i, \\
\hat{y} & = \text{MLP}(\hat{\mathbf{h}}),
\end{align}
where $\text{MLP}$ consists of a dropout layer, a fully connected layer with 2048 neurons, a ReLU activation, and a final fully connected layer that outputs a single neuron representing the predicted ECR score $\hat{y}$.

We optimize the model using the MSE loss between predictions and ground-truth labels:
\begin{equation}
\mathcal{L}_{\text{MSE}} = \frac{1}{N} \sum_{i=1}^{N} (\hat{y}_i - y_i)^2,
\end{equation}
where $\hat{y}_i$ and $y_i$ denote the predicted and ground-truth ECR scores for the $i$-th sample, respectively, and $N$ is the total number of training samples in a batch.

\subsubsection{Qwen2.5-VL}

Qwen2.5-VL is a general-purpose vision-language model capable of processing both visual and textual modalities. In our setting, it takes the same visual input $\bm{x}$ and textual metadata $T_{\text{title}}$, $T_{\text{description}}$ as the VideoLLaMA2 branch, excluding the audio modality. For memory efficiency, the visual input is resized to a maximum spatial resolution of $768 \times 28 \times 28$ before being encoded.

The keyframes $\bm{x}$ are first passed through the vision encoder $f_{\text{visual}}$ to extract a sequence of visual embeddings ${\mathbf{z}_i^{\text{visual}}}$ via Equation (\ref{visual_equation}). The multimodal prompt by embedding the visual tokens into a natural language template $P$ and appending the title and description as plain text:
\begin{quote}
\small
\texttt{The video frames: <frames tokens> ... <frames tokens>. How would you judge the engagement continuation rate of the given content, where engagement continuation rate represents the probability of watch time exceeding 5 seconds. The title of the video is: <title>, and the description of the video is: <description>.}
\end{quote}

The entire prompt $P$ is tokenized and processed by the autoregressive transformer decoder of Qwen2.5-VL. Unlike Video-LLaMA2, which employs a regression head to predict a scalar value, Qwen2.5-VL generates the ECR score directly in the form of next-token generation. Let the ground-truth score token sequence be denoted as $\mathcal{Y} = {y_1, \dots, y_T}$, where each $y_t$ is a token corresponding to the characters in the numerical score. The model is trained via standard cross-entropy loss:

\begin{equation}
\mathcal{L}_{\text{CE}} = - \sum_{t=1}^{T} \log P(y_t \mid y_{<t}, P).
\end{equation}

This formulation enables Qwen2.5-VL to model the ECR prediction task as a sequence generation problem, leveraging its strong language modeling capabilities to infer engagement scores in a token-wise manner.

\section{Experiments}

\subsection{Implementation Details}
We adopt \texttt{VideoLLaMA2.1-7B-AV} and \texttt{Qwen2.5-VL-7B-Instruct} as the backbone weights for the Video-LLaMA2 and Qwen2.5-VL models, respectively. For VideoLLaMA2, we freeze the vision encoder and fine-tune the remaining components. Training is conducted on two A800 GPUs with a batch size of 12 for one epoch, using the AdamW optimizer with a learning rate of $5 \times 10^{-5}$. In contrast, for Qwen2.5-VL-7B, we perform full-parameter fine-tuning, updating both the vision and language components. This training is performed on eight A800 GPUs with a batch size of 16 for one epoch. The AdamW optimizer is used with separate learning rates: $2 \times 10^{-6}$ for the vision encoder and $1 \times 10^{-5}$ for all other components.

\begin{table}[t]
\renewcommand{\arraystretch}{1.2}
\centering
\small
\caption{Performance comparison of different methods on the SnapUGC test set.}
\label{tab:main_results}
\begin{tabular}{lccc}
\hline
\textbf{Model} & \textbf{SROCC} & \textbf{PLCC} & \textbf{Final Score} \\
\hline
MinimalisticVQA~\cite{sun2024analysis} & 0.587 & 0.582 & 0.585 \\
Li24~\cite{li2024delving} & 0.657 & 0.665 & 0.660 \\
Qwen2.5-VL~\cite{bai2025qwen2}      & 0.665& 0.662& 0.664 \\
VideoLLaMA2~\cite{cheng2024videollama} & 0.691 & 0.701& 0.695 \\
\hline
\end{tabular}
\end{table}

\subsection{Validation Dataset}

We conduct experiments on the SnapUGC dataset~\cite{li2024delving}, which comprises short-form videos ranging from 5 to 60 seconds in duration. Each video is accompanied by user-provided metadata, including a title and description. The dataset is divided into three splits: 106,192 videos for training, 6,000 for validation, and 8,459 for testing. The provided ground-truth metric for evaluating video engagement is the Engagement Continuation Rate (ECR), which reflects the likelihood that a viewer watches a video for more than five seconds.

\subsection{Evaluation Metrics}
The performance of the video engagement prediction model is evaluated by comparing its predicted ECR scores against the ground-truth values using two correlation-based metrics. The Spearman Rank-Order Correlation Coefficient (SROCC) assesses the model's ability to preserve the relative ranking of videos in terms of engagement, while the Pearson Linear Correlation Coefficient (PLCC) measures the linear agreement between predicted and actual ECR scores. The final performance score is computed as a weighted combination of the two metrics:
\[
\text{Final Score} = 0.6 \times \text{SROCC} + 0.4 \times \text{PLCC}.
\]

This composite metric provides a balanced evaluation of both ranking consistency and prediction accuracy for engagement estimation.

\begin{table}[t]
\renewcommand{\arraystretch}{1.2}
\centering
\caption{Performance comparison of VideoLLaMA2 with different numbers of input frames on the SnapUGC test set.}
\label{tab:ablation_frames_results}
\resizebox{0.48\textwidth}{!}{
\begin{tabular}{lcccc}
\hline
\textbf{Model} & \textbf{\# Frames} & \textbf{SROCC} & \textbf{PLCC} & \textbf{Final Score} \\
\hline
VideoLLaMA2 & 5 & 0.686 &  0.695& 0.690 \\
VideoLLaMA2 & 8 & 0.691 & 0.701& 0.695 \\
\hline
\end{tabular}
}
\end{table}

\begin{table}[t]
\renewcommand{\arraystretch}{1.2}
\centering
\caption{Performance comparison of VideoLLaMA2 variants on the SnapUGC test set under different proportions of training data.}
\label{tab:training_proportion_results}
\resizebox{0.48\textwidth}{!}{
\begin{tabular}{lcccc}
\hline
\textbf{Model} & \textbf{Training Proportion} & \textbf{SROCC} & \textbf{PLCC} & \textbf{Final Score} \\
\hline
VideoLLaMA2 & 60\% & 0.687 & 0.696 & 0.691 \\
VideoLLaMA2 & 100\% & 0.691 & 0.701 & 0.695 \\
\hline
\end{tabular}
}
\end{table}

\subsection{Compared Methods}
We compare our methods with two representative baselines: a video quality assessment method, MinimalisticVQA~\cite{sun2024analysis}, and a video engagement prediction method, Li24~\cite{li2024delving}. MinimalisticVQA performs end-to-end training of a spatial quality analyzer along with a pre-trained temporal quality analyzer to assess visual quality. Li24 is a two-stage video engagement prediction framework that first extracts a diverse set of multimodal features, including semantic content, distortion-aware features, captioning features, sound classification features, etc., and then employs a MLP for engagement score regression. For fair comparison, both baseline methods are re-trained on the SnapUGC dataset~\cite{li2024delving}.

\subsection{Performance Comparison}

Table~\ref{tab:main_results} presents the performance of our models and the baseline methods on the SnapUGC test set. As shown in Table~\ref{tab:main_results}, MinimalisticVQA perform poorly on SnapUGC, suggesting that pure VQA models struggle to capture video engagement. In contrast, Qwen2.5-VL slightly outperforms Li24, a method that leverages rich engagement-related features, demonstrating that LMMs can automatically learn comparable engagement representations without relying on predefined features. Furthermore, VideoLLaMA2 significantly outperforms Qwen2.5-VL, despite the latter being a more recent LMM with strong general visual understanding capabilities. We attribute this improvement to the inclusion of background audio in VideoLLaMA2, highlighting the importance of audio cues in modeling viewer engagement.

\subsection{Ablation Studies}
We conduct three ablation studies to investigate the impact of (1) the number of input video frames, (2) the amount of training data, and (3) the choice of regression strategy on model performance.

\noindent\textbf{Number of Video Frames.}
To evaluate the effect of input frame count, we perform an ablation study on the number of keyframes used in VideoLLaMA2. As shown in Table~\ref{tab:ablation_frames_results}, increasing the number of keyframes consistently improves performance across all evaluation metrics. This suggests that incorporating a richer sequence of visual information enables the model to better capture temporal patterns and viewer engagement cues in UGC videos. The findings underscore the importance of frame sampling density in enhancing engagement prediction accuracy.

\noindent\textbf{Amount of Training Data.}
We analyze the impact of training data volume by comparing the performance of VideoLLaMA2 trained with 60\% and 100\% of the available data. As shown in Table~\ref{tab:training_proportion_results}, using the full training set results in consistently better performance across all metrics. This demonstrates that larger-scale supervision provides richer learning signals, allowing the model to better generalize engagement patterns. The results highlight the importance of data scale in training LMMs for video engagement prediction.

\begin{table}[t]
\renewcommand{\arraystretch}{1.2}
\centering
\footnotesize
\caption{Performance comparison of Qwen2.5-VL with different regression methods on the SnapUGC test set. ``Feature-based" refers to applying a two-layer MLP regressor on the hidden representations to predict engagement scores, optimized with a mean squared error loss. and ``Token-based" denotes directly generating the quality score as output tokens from the LMM, with training guided by a cross-entropy loss.}
\label{tab:ablation_regression_results}
\begin{tabular}{lcccc}
\hline
\textbf{Model} & \textbf{Regression} & \textbf{SROCC} & \textbf{PLCC} & \textbf{Final Score} \\
\hline
Qwen2.5-VL & Token-based & 0.665  &  0.662& 0.664 \\
Qwen2.5-VL & Feature-based & 0.674 & 0.679& 0.676 \\
\hline
\end{tabular}
\end{table}

\noindent\textbf{Regression Strategy.} We adopt different regression strategies for VideoLLaMA2 and Qwen2.5-VL. Specifically, VideoLLaMA2 employs a feature-based regression approach, where a two-layer MLP is used to map hidden features to quality scores, optimized with mean squared error (MSE) loss. In contrast, Qwen2.5-VL utilizes a token-based regression approach, where the model directly generates quality scores as output tokens and is trained using cross-entropy loss. To evaluate the effectiveness of these two strategies, we apply both regression methods to Qwen2.5-VL and report the experimental results in Table~\ref{tab:ablation_regression_results}. As shown in Table~\ref{tab:ablation_regression_results}, feature-based regression consistently outperforms token-based regression, suggesting that regressing the hidden features via an MLP is a more effective regression method for video engagement prediction. This advantage stems from the ability of feature-based regression to exploit continuous latent representations and optimize a smooth regression objective, which is more effective for fine-grained score prediction than the discrete and often noisy supervision provided by token-level cross-entropy loss.

\subsection{Performance on the EVQA Challenge}

\begin{table}[t]
\renewcommand{\arraystretch}{1}
\centering
\caption{Result of VQualA 2025 EVQA-SnapUGC Challenge.}
\label{tab:challenge_results}
\resizebox{0.48\textwidth}{!}{
\begin{tabular}{lccc}
\hline
\textbf{Team name} & \textbf{Final Score} & \textbf{SROCC} & \textbf{PLCC} \\
\hline
ECNU-SJTU VQA (ours) & \textbf{0.710} & \textbf{0.707} & \textbf{0.714} \\
IMCL-DAMO & 0.698 & 0.696 & 0.702 \\
HKUST-Cardiff-MI-BAAI & 0.680 & 0.677 & 0.684 \\
MCCE & 0.667 & 0.666 & 0.668 \\
EasyVQA & 0.667 & 0.664 & 0.671 \\
Rochester & 0.449 & 0.405 & 0.515 \\
brucelyu & 0.441 & 0.439 & 0.444 \\
\hline
\end{tabular}
}
\end{table}

Table~\ref{tab:challenge_results} summarizes the results of the ICCV 2025 VQualA EVQA-SnapUGC: Engagement Prediction for Short Videos Challenge. Our method, which ensembles four models, including three VideoLLaMA2 variants (from Table~\ref{tab:ablation_frames_results} and Table~\ref{tab:training_proportion_results}) and one Qwen2.5-VL model (from Table~\ref{tab:main_results}) achieves the highest final score of 0.714 and ranks first in the competition. This result demonstrates the effectiveness of our ensemble strategy and highlights the strong performance of our models in video engagement prediction.

\section{Conclusion}

In this work, we explore the effectiveness of large multimodal models for video engagement prediction in user-generated content. We evaluate two representative LMMs, VideoLLaMA2, which incorporates audio, visual, and textual modalities, and Qwen2.5-VL, which relies on visual and textual inputs. Experimental results on the SnapUGC dataset demonstrate that both models achieve competitive performance compared to state-of-the-art baselines, validating the potential of LMMs in modeling complex engagement patterns. Furthermore, the superior performance of VideoLLaMA2 over Qwen2.5-VL highlights the critical role of audio features in improving engagement prediction accuracy. By ensembling both models, our approach achieves first place in the ICCV VQualA 2025 EVQA-SnapUGC Challenge. We believe our findings offer valuable insights into the design of multimodal systems for real-world video engagement and user interaction modeling.

\section{Acknowledgement}
This work was supported in part by the National Natural Science Foundation of China under Grants 62301316 and 62371283.

{
    \small
    \bibliographystyle{ieeenat_fullname}
    \bibliography{main}
}

\end{document}